\documentclass[letterpaper, 10 pt, conference]{ieeeconf}

\IEEEoverridecommandlockouts                              	
\usepackage{booktabs}
\usepackage{cite}
\usepackage{amsmath,amssymb,amsfonts}
\usepackage{graphicx}
\usepackage[caption=false]{subfig}

\usepackage{algpseudocode}
\usepackage{textcomp}
 
\usepackage{hyperref}

 \newcounter{algorithm}
 \newcommand{\algcaption}[1]{%
   \par\noindent
   \refstepcounter{algorithm}%
   \textbf{Algorithm \thealgorithm:}
   #1\par
 }

\usepackage[normalem]{ulem}
\usepackage{xcolor}

\newcommand\subs[1]{_{\text{#1}}}

\algnewcommand{\LeftComment}[1]{\Statex \(\triangleright\) #1}

\def\BibTeX{{\rm B\kern-.05em{\sc i\kern-.025em b}\kern-.08em
    T\kern-.1667em\lower.7ex\hbox{E}\kern-.125em}}

\usepackage{aveasmargins}

\begin{document}

\title{\LARGE \bf
Automatic Odometry-Less OpenDRIVE \\Generation From Sparse Point Clouds
}

\author{Leon Eisemann$^{1}$ and Johannes Maucher$^{2}$
\thanks{*This publication was written in the context of the AVEAS research project (www.aveas.org), funded by the German Federal Ministry for Economic Affairs and Climate Action (BMWK) within the program ``New Vehicle and System Technologies''.}
\thanks{$^{1}$Leon Eisemann is with the department of Artificial Intelligence \& Big Data at
        Porsche Engineering Group GmbH, 71287 Weissach, Germany}%
\thanks{$^{2}$Johannes Maucher is with the Institute for Applied Artificial Intelligence at
Stuttgart Media University, 70569 Stuttgart, Germany}%
}

\maketitle
\thispagestyle{empty}
\pagestyle{empty}

\aveasSetMargins{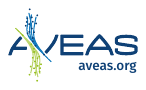}
\aveasSetIEEEFoot{2023}
\aveasSetIEEEHead{L. Eisemann and J. Maucher, ``Automatic Odometry-Less OpenDRIVE Generation From Sparse Point Clouds,'' 2023 IEEE 26th International Conference on Intelligent Transportation Systems (ITSC), Bilbao, Spain, 2023, pp. 681--688}{10.1109/ITSC57777.2023.10421842}

\begin{abstract}
High-resolution road representations are a key factor for the success of (highly) automated driving functions. These representations, for example, high-definition (HD) maps, contain accurate information on a multitude of factors, among others: road geometry, lane information, and traffic signs.
Through the growing complexity and functionality of automated driving functions, also the requirements on testing and evaluation grow continuously. This leads to an increasing interest in virtual test drives for evaluation purposes. As roads play a crucial role in traffic flow, accurate real-world representations are needed, especially when deriving realistic driving behavior data. 
This paper proposes a novel approach to generate realistic road representations based solely on point cloud information, independent of the LiDAR sensor, mounting position, and without the need for odometry data, multi-sensor fusion, machine learning, or highly-accurate calibration. As the primary use case is simulation, we use the OpenDRIVE format for evaluation.
\end{abstract}

\section{Introduction} \label{introduction}

The growing complexity of Advanced Driver Assistance Systems (ADAS) leads to higher requirements for testing and evaluation of these systems. Especially the functional safety and the safety of the intended functionality (SOTIF) of (highly) automated driving (AD) functions must be ensured in a large number of traffic situations. One possible method for the safety assessment is real-world driving on public roads, with statistical validation\cite{karunakaran_efficient_2020}. The increasing complexity implies the need for more and more kilometers driven for a comprehensive evaluation. In \cite{wachenfeld2016}, a statistical analysis shows that validation by real test drives is infeasible for future automated functions. Based on the average driven kilometers between two fatal accidents on German highways, the authors derive that 6.61 billion kilometers are needed to encounter at least one similar situation.

With these prerequisites, the adoption of simulation is critical to reduce development time and ensure system robustness and reliability\cite{park-creating-2020, xinxin-csg-2020}.

For the extraction and simulation of a traffic scenario, the authors in \cite{ xinxin-csg-2020} identify three critical elements: a road representation, the traffic participants, and the environment. A concept of the road is hereby considered an integral part, as the structure and course of the road are one of the key influential factors in the behavior of traffic participants. Albeit the importance of road networks for evaluating driving functions, their creation is still an open research topic.

In the field of simulation, extraction methods of respective representations can be divided into four categories\cite{bao-high-definition-2022}:

First, high-definition (HD) maps or aerial images typically provide information about the road network at a centimeter-level accuracy\cite{park-creating-2020}. This information can then be transformed into the respective format for simulation, as shown in \cite{park-creating-2020}.

Second, the extraction and evaluation of road structures from a mobile mapping platform. Mobile mapping platforms are vehicles fitted with a high number of sensors, including Global Navigation Satellite System (GNSS), inertial measurement unit (IMU), and light detection and ranging (LiDAR), which are optimized to collect detailed environment representations\cite{barsi-role-2020}\cite{bao-high-definition-2022}. 

\begin{figure}[t]
\centering
\includegraphics[width=1\columnwidth]{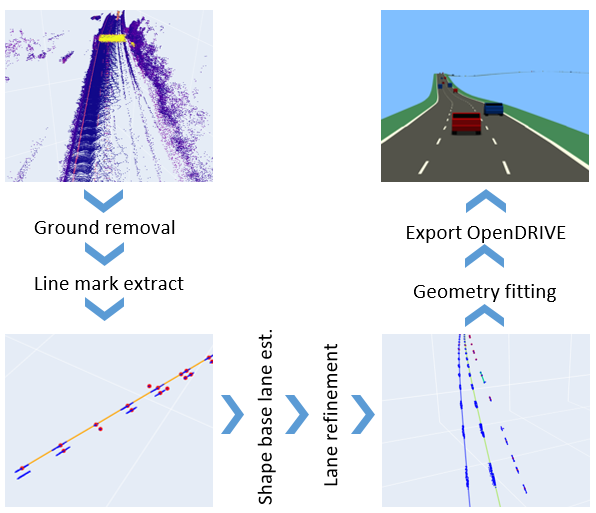}
\caption{General overview of our proposed method. We first build an overall point cloud of the road, filter the respective line markings, generate the respective lanes based on the markings, and in the last step, export into OpenDRIVE. Hereby, the proposed method solely uses point cloud information and exploits geometric properties of lane markings.}
\label{fig:process-overview}
\end{figure}

Third, the fusion of mobile mapping approaches and external data sources is common to further reduce shortcomings of individual methodologies\cite{bao-high-definition-2022}. As shown in \cite{ma-boundarynet-2022}, the authors thereby overcome the occlusion of LiDAR point clouds by fusing the resulting road boundaries with satellite imagery.

Finally, maps created by service providers. Especially in automotive simulation, these are commonly leveraged, for example, in \cite{montanari-maneuver-based-2021, richter-systematische-2016, schwab-spatio-semantic-2020}. However, since these service providers operate commercially, there is hardly any publicly available information about their methods and degree of automation. Furthermore, these come with additional costs, coordination, and planning efforts. For this reason, they are considered out of scope in the context of this work.

\begin{table*}[ht]
\caption{Comparison of the prerequisites of previous research and the resulting length of reconstruction. If not explicitly stated, it is derived from the evaluation and graphics shown. In comparison, our methodology needs the fewest prerequisites in terms of sensors used and mounting positions, can reconstruct other traffic participants, and achieves the longest reconstruction.}
\centering
\resizebox{2\columnwidth}{!}{%
\begin{tabular}{@{}lcccccr@{}}
\toprule
\textbf{Authors}            & \textbf{nr. sensors} & \textbf{hor. FoV} & \textbf{mounting pos.} & \textbf{rec. traffic} & \textbf{odometry needed} & \textbf{reconstruction length} \\ \midrule
Haala et al.\cite{haala-mobile-2008}       & 4                    & 360°           & mult. + ground facing                     & no                       & GNSS + IMU               & 1,500 m                             \\ \midrule
Ibrahim et al.\cite{ibrahim-high-2021}     & 1                     & 360°           & top                              & no                       & no                        & 3,600 m                             \\ \midrule
Yang et al.\cite{yang-semi-automated-2013}       & 2                    & 360°           & mult.                               & no                       & no                        & 2,200 m                             \\ \midrule
Chiang et al.\cite{chiang_automated_2022}      & 1                     & 360°           & ground facing                    & no                       & GNSS + IMU               & 400 m                              \\ \midrule
Karunakaran et al.\cite{karunakaran-automatic-2022} & 7                    & 360°           & mult. + ground facing             & yes                        & IMU                      & 200 m                              \\ \midrule
\textbf{Ours}               & \textbf{1}           & \textbf{82°}   & \textbf{front}                    & \textbf{yes}               & \textbf{opt.}            & \textbf{14,000 m}                    \\ \bottomrule
\end{tabular}%
}
\label{tab:sota-req-comparison}
\vspace{0.2cm}
\end{table*}

The requirements for the creation of road representations depend on the use case. We deemed none of the above-mentioned feasible for use in automotive simulation. 
In application, map or aerial-based methods have been shown to be inaccurate and often not up to date, especially in the case of road works.
Road works have special importance for the safety assessment of automated driving functions, as \cite{yang-workzone-trafficaccidents} has shown that the presence of a work zone is likely to increase the crash rate.

Therefore, the creation of road models through a mobile mapping platform is more flexible. 
However, this flexibility often depends on very specific sensor setups and mounting points. This leads to high costs for equipment, sensors, and workforce and hinders the broad availability of these platforms.
Also, these vehicles are often unable to record traffic participants, resulting in two separate recording drives or vehicles. In the case of additional drives, there is no guarantee that the course of the road is unchanged.
Furthermore, recent research \cite{ibrahim-high-2021, yang-semi-automated-2013, chiang_automated_2022, karunakaran-automatic-2022} has shown to be reliant on specific sensor types, e.g., line-wise scanners, sensor characteristics, like 360° field of view (FoV), or mounting points, which makes the adoption on other data sources complex.

To solve some of the mentioned disadvantages, we propose a generic approach capable of producing a standardized map format. This is achieved in an automated manner, without using odometry or any prior knowledge about the road. We use a sparse LiDAR point cloud derived directly from real-world test drives of cut-in scenarios.

In the evaluation, we compare our results quantitatively to other map sources and show the quality through a trajectory-based simulation of the original test drive. As a map standard, we adopt OpenDRIVE, which is further explained in Sec.~\ref{sec:mapformat}, where we contribute to the creation from real-world measurement data. We propose a distance-based look ahead / back geometry fitting to solve geometric leap and kink problems described in \cite{schwab-validation-2022}.

The paper is structured as follows: Sec.~\ref{sec:background} provides the motivation and a brief overview of prior research. 
Sec.~\ref{sec:method} presents the data used and applies the proposed method to highway scenarios. Finally, Sec.~\ref{sec:results-discussion} evaluates the results and gives an outlook on limitations and avenues for future research.

\section{Background}\label{sec:background}

The following section reviews 3D point cloud mapping approaches and introduces the map standard used in this work.

\subsection{Related Work}
For constructing the road networks of the city center of Stuttgart, Germany, with an area of 1.5 km x 2 km, the presented study in \cite{haala-mobile-2008} uses a mobile mapping platform with 360° scanner FoV and GNSS. The results show the possibility to create a point cloud-based representation of city areas with high accuracy.

In \cite{ibrahim-high-2021}, data is collected by driving circles around city blocks. The LiDAR scans of each block are then processed, including subsampling, noise removal, deduplication, and smoothing. This approach creates a merged point cloud for the Central Business District of Perth, Australia.

Leveraging curb points, \cite{yang-semi-automated-2013} proposes the extraction of road models based on point clouds from a 360° FoV multi-LiDAR setup. In the described process, a sliding window approach is adapted to filter non-ground points and curb points in every scan line. Individual points are tracked and refined, allowing global consistency to be preserved.

An automated approach exporting OpenDRIVE is presented in \cite{chiang_automated_2022}, using the trajectory of the ego vehicle and a ground-facing laser scanner. A multistage reconstruction process is used to recreate the road network based on extracted road edges, line marking classification, and the trajectory. Compared to other methods, the authors can also classify road segments into lane lines, stop lines, arrows, and others.

The methods mentioned above focus on the sole generation of road representations. As stated in Sec.~\ref{introduction}, this would result in separate recordings for the road and traffic participants, which yields several disadvantages. First, referencing the same map coordinate system while compensating for the inaccuracy of GNSS and tracking algorithms or systems. Second, based on the time difference between recordings, the course of the road can be changed, e.g., through construction sides or accidents. These two main factors render these approaches labor-intensive and uncertain for reliable usage in AD simulation or traffic behavior analysis.

A higher focus on traffic behaviour is shown in \cite{karunakaran-automatic-2022}. In addition to a road model, the authors are able to derive and analyse traffic participants. The authors propose a fully automated point cloud-based system generating OpenDRIVE and OpenSCENARIO \cite{asam-openscenario} from a single recording. Data is sourced from a vehicle fitted with six LiDAR sensors and an additional LiDAR facing the ground. Based on tracking information, they separate between dynamic and static objects, sort the static point cloud in reference to the ego vehicles odometry. 
Through the exploitation of the line-wise LiDAR scanning pattern, lane markings are filtered. These are then transformed into the Lanelet2 \cite{poggenhans2018lanelet2} format, which is again converted into OpenDRIVE. Although this work focuses on the driving scenario extraction and the resulting road is not evaluated, it can be assumed that they can reconstruct several hundred meters based on the extracted scenarios.

The preconditions of the presented methods and ours are compared in Tab.~\ref{tab:sota-req-comparison}. The need for 360° FoV, the reliance on special sensor characteristics, mounting positions, and highly accurate odometry information hinder the adaption of these methodologies without the concrete vehicle at hand. Our motivation is to achieve precise road models, without leveraging sensor specifications, a vast number of sensors, and without the need for highly-accurate calibration, multi-sensor fusion, or machine learning. In addition, we aim for direct creation from test drives and independence from odometry information. The later allows the use of different LiDAR sources, e.g., multiple vehicles, drones, and static sensors.

\subsection{Map Format}\label{sec:mapformat}
In \cite{chiang_automated_2022}, the authors extensively compare different formats for HD map creation, e.g., Autoware Vector Map, OpenDRIVE, Lanelet2, and Navigation Data Standard. Based on their evaluation and our focus on automotive simulation, OpenDRIVE is selected as HD map format. One of the main reasons is the wide adoption in the automotive industry and support by tools and software platforms for creating, editing, and using road data in simulation and testing scenarios \cite{chiang_automated_2022}. OpenDRIVE originates from the automotive industry's efforts for standardization and provides a comprehensive and flexible framework for describing the geometry, topology, and attributes of roads\cite{asam-opendrive}.

The standard defines the overall course of the road along a reference line. Further elements, e.g., elevation and lanes, are then attached to it. Per definition, a reference line can consist of four types of geometries, namely \textit{line}, \textit{spiral}, \textit{arc}, or \textit{paramPoly3} (parametric cubic curve). These geometries are usually concatenated to describe a more detailed or longer road structure\cite{asam-opendrive}.

\section{Method}\label{sec:method}
Similar to other approaches, we also adopt LiDAR data as input. Over the last few years, the costs for LiDAR sensors dropped significantly, which facilitates their use in research. They provide high positional accuracy of detections, and the measured reflectivity or intensity allows easier recognition of line markings. In contrast to previously mentioned methods, see Tab.~\ref{tab:sota-req-comparison}, we do not use a high-resolution LiDAR, a special mounting position, or a 360° FoV, and instead, use a single front-mounted LiDAR sensor with an 82° horizontal FoV. In the following section, we present our proposed method. For better comprehension, individual steps are summarized as pseudocode, and functions are marked in \textit{italics} in the text.

\subsection{Data \& Preprocessing}
As data collection vehicle, we make use of the JUPITER platform as described in \cite{Haselberger-Jupiter}. The platform consists of a Porsche Cayenne fitted with multiple Livox LiDARs and GeneSys ADMA GNSS inertial system. To show our approach's robustness, we exclusively use the front-mounted LiDAR. 
This sensor features a complex inhomogeneous scanning pattern, which shows our approach's independence of sensor characteristics. 
Compared to other methods, the data collection was not focused on road generation but on the acquisition of dynamic driving behavior and critical driving scenarios as part of the AVEAS~\footnote{\url{www.aveas.org}} \cite{aveas-paper-2023} research project. 
To derive realistic traffic behavior, data was collected mainly on the German interstate highways A8 and A81 around Stuttgart at peak traffic times. The speed profile of the recordings ranges from 80 to 140 km/h in the considered parts.
We extract the data between the highway's entry and exit, with the entry as the origin of our world coordinate system. The world coordinate system is defined in the 3D Euclidean space and is axis-aligned with the geodesic coordinates. 

We convert each LiDAR scan into a point cloud with \textit{X,~Y,~Z,~Reflectivity} for every detection. Resulting point clouds are stored in the vehicle coordinate system as defined in \cite{iso-8855}. The corresponding GNSS coordinates are stored separately.

\begin{figure}[t]
\vspace{0.1cm}
\algcaption{Calculation of Candidate Lines}\label{alg:calc-candid-lines}
  
  \begin{algorithmic}[1]
  
    	\State \textit{clusters.append(\textbf{DBSCAN}(point cloud))}
    	
    	\If{\textit{cluster.length} $\ge$ \textit{threshold}} \label{alg:splitclusters}
    		\State \textit{clusters.append(\textbf{SplitCluster}(cluster))}
	\EndIf

        \State \textit{cCenters.append(\textbf{CalcCenterBB}(cluster))} \Comment{parallel}
    	\State \textit{dirVecs.append(\textbf{LineRANSAC}(cluster))} \Comment{parallel}
    	\State \textit{dirVecs $=$ \textbf{UnifyDirection}(dirVecs))} \Comment{parallel}
    	\For {\textit{cluster} in \textit{clusters}}
    		\If{\textit{currentLine} $\ge$ 2}
    			\State \textit{dirVec} $=$ Eq.~\eqref{eq:dir-vec-stabilization}
		\EndIf
		\State \textit{candidates.append(\textbf{SearchMark}(cCenters, dirVec))}
	\EndFor
		
	\While {$connection$}
		\State \textit{candid} $\gets$ \textit{candidates}
    		\If{\textit{\textbf{Intersec}}(\textit{candid.end}, \textit{candidates.start})}
    			\State \textit{candid} $=$ \textit{\textbf{Combine}}(\textit{candid}, \textit{intersec})
    			\State \textit{candid} $=$ \textit{\textbf{DistSort}}(\textit{candid})
    			\State \textit{candidates.append(candid)}
    			\State \textit{connection} $=$ \texttt{True}
		\EndIf
	\EndWhile
	\State \Return \textit{candidates}
	\LeftComment{Overview over the lane generation process.
	\textit{\textbf{Function}} names are used for visibility and are further explained in Sec.~\ref{sec:gen-candid-lines} and Sec.~\ref{subsec:comb-candid-lines}}	
  \end{algorithmic}
\end{figure}

\subsection{Line Marking Extraction} \label{subsec:line-marker-extrac}
Since the following steps of our method use point cloud data of line markings as a representation, we showcase the extraction of these. 

As data source, we use the LiDAR point clouds directly in the vehicle coordinate system for ground plane estimation. For outlier reduction, the point clouds are first cropped to the maximum sensor range specified by the manufacturer. From this, only points below the mounting position of the sensor are considered. This reduces the number of points originating from surroundings, e.g., traffic signs or bridges, and allows a more robust ground plane estimation. Then, through a \textit{RANSAC}\cite{ransac-fischler}-based approach, we estimate the ground plane and artificially raise it by a margin of 15 cm to account for possible road normal curvatures. The value of 15~cm is hereby chosen, as it corresponds to the average height of curbstones in Europe. Lane markings are filtered through their high reflectivity in the LiDAR returns. Based on evaluations on our dataset, we choose an reflectivity of 50 \% as filter threshold. While the measured reflectivity of the sensor under use is estimated through a Lambertian reflectance model, the interpretation and calculation depend on the sensor used, and likely would have to be adjusted for new data.

The extracted lane markings are accumulated and translated to their respective position in the world coordinate system, based on GNSS as shown in Fig.~\ref{fig:process-overview}. 
Although we use the GNSS-derived translation, one could also derive the translation based on, e.g., a SLAM- or ICP-based approach as in \cite{livox-slam-2023,vizzo2022arxiv}.
Independently from the translation, reflectivity-based filtering leads to considerable outliers through multipath reflections and other environmental impacts. Therefore, an additional radius-based outlier reduction is done. Based on different evaluations on our data, we remove each point with less than four neighboring points in a half-meter radius.

\subsection{Generation of Lane Marking}\label{sec:gen-candid-lines}

\begin{figure}[tb]
\centering
\vspace{0.02in}
\includegraphics[width=0.995\columnwidth]{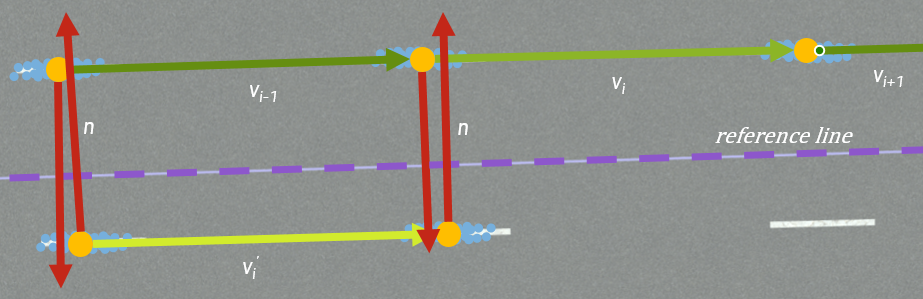}
\caption{Overview over lane generation and positional calculation. Lane marking point clouds are displayed in \textit{blue}, and respective estimated centers in \textit{orange}. The directional vectors calculated by Eq.~\ref{eq:dir-vec-stabilization} are shown in shades of \textit{green} and derived normal-vectors in \textit{red}. The resulting \textit{reference line} is displayed in \textit{purple}.}
\label{fig:method-overview-lanegen}
\end{figure}

The overall generation of candidate lane markings is listed in Alg.~\ref{alg:calc-candid-lines} for better comprehension.

As our method depends on the individual lane markings' geometry, we first cluster points that belong to each lane mark. Due to the fact, that the total number of markings is unknown at this stage of the process, we adopt \textit{DBSCAN} \cite{dbscan-ester} clustering.

Analyses of the resulting clusters indicate that depending on the ego position on the road, clusters of solid lane markings can reach up to more than 1 km in length. These continuous markings are valuable for lane generation since they contain dense information about the course of the road, also in-between dashed markings. To integrate this information into the further process and achieve a unified method, markings reaching over 30 m in length are divided into uniformly sized slices; see Alg.~\ref{alg:calc-candid-lines} line~\ref{alg:splitclusters}. The resulting slices are 6 meters long, which equals the standard size of dashed markings on German highways \cite{frank-leitfaden-2017}.

The reconstruction of lane geometries in point cloud data is a non-trivial task. Through the nature of point clouds, methods like nearest neighbor would yield wrong connections between lanes.
Hence, for the construction of the lanes, related research, e.g., \cite{karunakaran-automatic-2022, chiang_automated_2022}, integrates ego trajectory information to solve this limitation, which in return requires an accurate GNSS and/or IMU setup.

To be able to reconstruct the lanes without the use of ego information, we exploit the geometric properties of the individual lane markings. The underlying assumption is that clusters derived from dashed as well as solid lane markings are elongated in the direction of the road\cite{frank-leitfaden-2017}. 
Therefrom, we evaluate the direction of each cluster through a \textit{RANSAC}-based line calculation, see \textit{LineRANSAC}, to derive a directional vector $\hat{v}^*$ for each line marking. Since our approach deliberately avoids the direct utilization of a driving direction, and RANSAC may return both positive or negative directional vectors for the symmetric lane marks, we subsequently
 unify them based on the directional vector between the origin of the coordinate system and the cluster centers in Alg.~\ref{alg:calc-candid-lines} as \textit{UnifyDirection}.

To achieve a stable connection between markings, calculations leverage full 3D point cloud information. In addition, this also reduces the chances of interferences through high-reflectivity outliers still contained in the point cloud and, thus, also lowers the quality requirements of the previous steps.
A search along the directional vector is executed from the cluster closest to the origin.

Since the directional vector $\hat{v}^*$ can contain a high variation and inaccuracies through partial scans or alignment issues, we calculate any further vector $\hat{v}$ based on Eq.~\eqref{eq:dir-vec-stabilization} with $\gamma = 0.5$, to provide additional stability:
\begin{equation}
\hat{v}\subs{i+1} = \gamma \cdot \hat{v}^*\subs{i} + (1-\gamma) \cdot \hat{v}^*\subs{i-1}
\label{eq:dir-vec-stabilization}
\end{equation}

As shown in Alg.~\ref{alg:calc-candid-lines} and Fig.~\ref{fig:method-overview-lanegen}, \textit{SearchMark} iteratively calculates the next expected mark center $p\subs{c}$ based on the distance $d$, through $p\subs{c} = \hat{v} \cdot d$, where a ball radius search for the next center point is done. The overall search is executed in three-meter steps for a total length of 27 m, which is 1.5$\times$distance between the center of each line marking on German highways \cite{frank-leitfaden-2017}. Through the high upper limit and step-wise procedure, we expect our method to work in other countries as well. The ball radius search is executed with a radius of 1.75 m, which was chosen as it equals half of the minimum lane width \cite{frank-leitfaden-2017}. If a point is found, it is added to the stack, and an additional point-to-line-segment calculation is done to add missed clusters. The process is repeated as long as cluster centers are found for the current line. Resulting \textit{candidate} lines are assigned a random ID for further processing.

\begin{figure}[tb]
  \vspace{0.1cm}
	\algcaption{Calculation of reference line}\label{alg:calc-ref-line}
  \begin{algorithmic}[1]
    	\State \textit{normVecs} $=$ \textit{\textbf{CalcNormVec}}(\textit{candidates})  
    	\For {\textit{normVec} in \textit{normVecs}}
    		\If{\textit{\textbf{Intersec}}(\textit{normVec}, \textit{candidates})}
    			\State \textit{dist} $=$ \textit{\textbf{dist}}(\textit{normVec.start}, \textit{intersec.point})
    			\State \textit{dist} $=$ \textit{dist} \% \textit{3.0m}
    			\State \textit{lookupRel.append(lane{ID}, intersecSide, dist)}
		\EndIf
		\State \textit{relativeLookups} $\gets$ \textit{lookupRel}
    \EndFor
		
	\For{ \textit{lookup} in \textit{relativeLookups}}
    		\If{\textit{lookup} in \textit{globalLookup}}
    			\State \textit{globalLookup.append(lookup.allIDs)}
		\EndIf
	\EndFor
	\State \Return \textit{globalLookup}
\LeftComment{Overview over the lane generation process. \textit{\textbf{Function}} names are used for visibility and explained in Sec.~\ref{subsec:calc-ref-line}}
  \end{algorithmic}
\end{figure}

\subsection{Combining Candidate lines}\label{subsec:comb-candid-lines}
Since outliers are reduced through the generation of the candidates, higher tolerances can be used for the combination of candidates. 
As the correct order of the markings is needed, the previously generated candidate lines are sorted by closest point sampling starting from the origin, in Alg.~\ref{alg:calc-candid-lines} as \textit{DistSort}. On this basis, another point-to-line-segment check is executed to combine candidates which starting or end point is located between another candidate segment. 
An additional search is done to achieve a more stable connection between the candidates and higher stability against occlusion, similar to the previous steps. This is done from the end point of every candidate for a total of 63 meters. If a matching candidate is found, the search candidate's ID is assigned by \textit{Combine}, and the resulting line is sorted again. This whole process repeats until no new combination is found.

\subsection{Calculation of Reference Line}\label{subsec:calc-ref-line}
\begin{figure}[!tbp]
\centering
\vspace{0.02in}
\includegraphics[width=0.75\columnwidth]{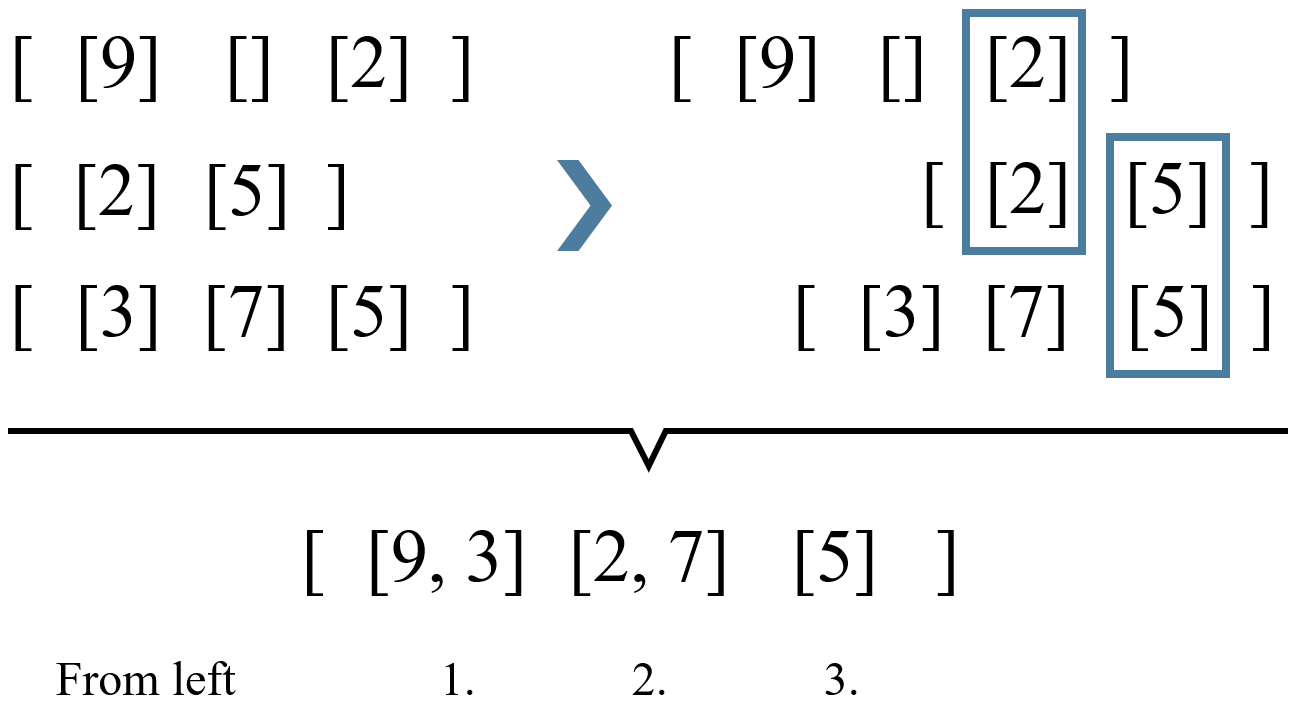}
\caption{Calculation of the \textit{GlobalLookup}, as described in Sec.~\ref{subsec:calc-ref-line}. The relative position is sorted through line ids in the relative positional lookups.}
\label{lookup-candid-ids}
\end{figure}

Since our method does not rely on the ego vehicle's odometry, we estimate the number of lanes based on the relative position between lane markings. These relative lookups are then solved to a global relation, as shown in Alg.~\ref{alg:calc-ref-line}.

In the first step, we transfer the calculated lanes into 2D space by simply omitting the \textit{Z}-dimension for the calculation. To evaluate the position of the lanes relative to each other, we first calculate the normal vectors $n$ for each line segment between cluster centers. Through leveraging the directional unification, we calculate the normals \textit{left} and \textit{right} of the segments in \textit{CalcNormVec}. The relative position of each lane is then estimated by the line intersection between the normals and every segment of candidates, as shown in Fig.~\ref{fig:method-overview-lanegen}. Additionally, the distance of the intersection towards the candidate line segment is evaluated and used for estimating the lane widths based on governmental regulations. 
For each lane, the resulting intersections and distances are saved in \textit{RelativeLookup}, as described in Alg.~\ref{alg:calc-ref-line}. 
These lookups are combined based on similar line ids into a \textit{GlobalLookup}, as shown in Fig.~\ref{lookup-candid-ids}. For example, if a lane with ID 3 has ID 5 to the right and another lane with ID 7 is intersected left from ID 5, then it can be concluded that both lane ID 3 and ID 7 are partial scans of the same lane.

In doing so, we achieve what we call \textit{superlines}, which contain each candidate's global positional lane offset. 

This process yields the number and width of lanes, from which we calculate the road's centerline. Based on the width of the road and the known relative offset of each lane, the normal vector towards the center of the road is calculated, and a new point is created with \textit{X, Y}-coordinates. To keep the height information of the road, the original \textit{Z}-information of the candidate lines is added. This process is done for every line marking center in each \textit{superline}, which results in a high resolution \textit{reference line}, see \textit{refLine}.

\subsection{Export of OpenDRIVE}\label{subsec:export-opendrive}

\begin{figure}[tb]
  \vspace{0.1cm}
	\algcaption{Calculation of OpenDRIVE}\label{alg:export-opendrive}
  \begin{algorithmic}[1]
    	\State \textit{segments} $=$ \textit{\textbf{SplitByDist}}(\textit{refLine})
    	\For {\textit{seg} in \textit{segments}}
    			\State \textit{geo.segRotation} $=$ \textit{\textbf{EvalRot}}(\textit{seg})
    			\State \textit{lookAhead} $=$  \textit{\textbf{CalcDist}}(\textit{seg.end}, \textit{refLine})
    			\State \textit{lookBack} $=$ \textit{\textbf{CalcDist}}(\textit{seg.start}, \textit{refLine})
    			\State \textit{seg} $=$ \textit{\textbf{CompensateRot}}(\textit{seg}, \textit{geo.segRotation})
    			\State \textit{p} $=$ \textit{\textbf{NormalizeDist}}(\textit{seg}, \textit{lookBack}, \textit{lookAhead})
    			\State \textit{geo.paramPolyU} $=$ \textit{\textbf{WeightedFit}}(\textit{p.x})
    			\State \textit{geo.paramPolyV} $=$ \textit{\textbf{WeightedFit}}(\textit{p.y})
    			\State \textit{geo.paramPolyZ} $=$ \textit{\textbf{Fit}}(\textit{seg.z})
    			\State add \textit{geo} to OpenDRIVE planView
	\EndFor

\LeftComment{Overview of the OpenDRIVE generation. \textit{\textbf{Function}} names are used for visibility and are further explained in Sec.~\ref{subsec:export-opendrive}}
  \end{algorithmic}
\end{figure}

As the re-calculation and direct export of measurement data into OpenDRIVE is a nontrivial task, it is outlined in more detail. The overall steps are displayed in Alg.~\ref{alg:export-opendrive}.

The OpenDRIVE standard recommends using parametric cubic curves to generate complex geometries from measurement data\cite{asam-opendrive}. 
The amount of information that can be displayed in one curve is limited by their cubic nature. To circumvent this limitation, we split the calculated continuous \textit{reference line} into 100 m long segments for the export. 
As presented in Alg.~\ref{alg:export-opendrive}, this is done through the \textit{SplitByDist} function. 
Important in this context is that the overall continuity of the road model has to be preserved, despite the segmentation. This is especially important for the usage in AD development. As described in \cite{schwab-validation-2022}, geometric leaps or kinks can cause vehicle dynamic models to fail. 
Primarily through offsets in the extracted lane marks and therefrom derived \textit{reference line}, geometric leaps and kinks are of particular interest.
 
To prevent these and allow a smooth transition between geometric elements, we introduce a distance based \textit{look ahead} and \textit{look back} index. These indexes contain parts of the predecessor, respectively successor segment, used for curve fitting but are not part of the final geometry. From different evaluations on our data, we defined the size for these lookups as 1/8 of the segment length.

Geometric elements consist of a local Cartesian coordinate system \textit{U, V}, which is placed in the inertial \textit{X, Y} coordinate system, see \cite{asam-opendrive}. For the correct placement in the inertial system, each element is rotated by a heading parameter. 
To estimate this parameter, the mean of all marking centers in the segment is evaluated, and the angle towards the \textit{X}-axis is derived. The line segment is then compensated by the negative heading to achieve the \textit{U, V} coordinate system for estimating the curve parameters.

Curve fitting for \textit{ParamPolyU} and \textit{ParamPolyV} integrates the \textit{look ahead} and \textit{look back} index to prevent the previously mentioned geometric kinks and leaps.

Additionally to the indexes, we use a weighting of the first and last original lane marking for the fitting of the \textit{U, V} cubic curves. Combined with the extended indexes, this prevents most geometric offsets, as through the \textit{look ahead} and \textit{back} indexes, the successor and predecessor shapes are integrated into the curves' fit, and the weighting prevents leaps.
Calculated parameters are then saved in the geometric object with the previously derived centered \textit{X, Y}-coordinates and heading.

Road elevation is defined as an additional element along the calculated road geometry and is calculated directly from the original reference markers, omitting weighting and \textit{look ahead} and \textit{look back} indexes.

The calculation of  super-elevation is done in the same manner. The rotation of the road surface is hereby derived from the calculated ground plane in  Sec.~\ref{subsec:line-marker-extrac}.

\begin{figure*}[!tp]
	\centering
	\subfloat[1][\label{fig:qual-examples-1}]
	{\includegraphics[width=0.675\columnwidth]{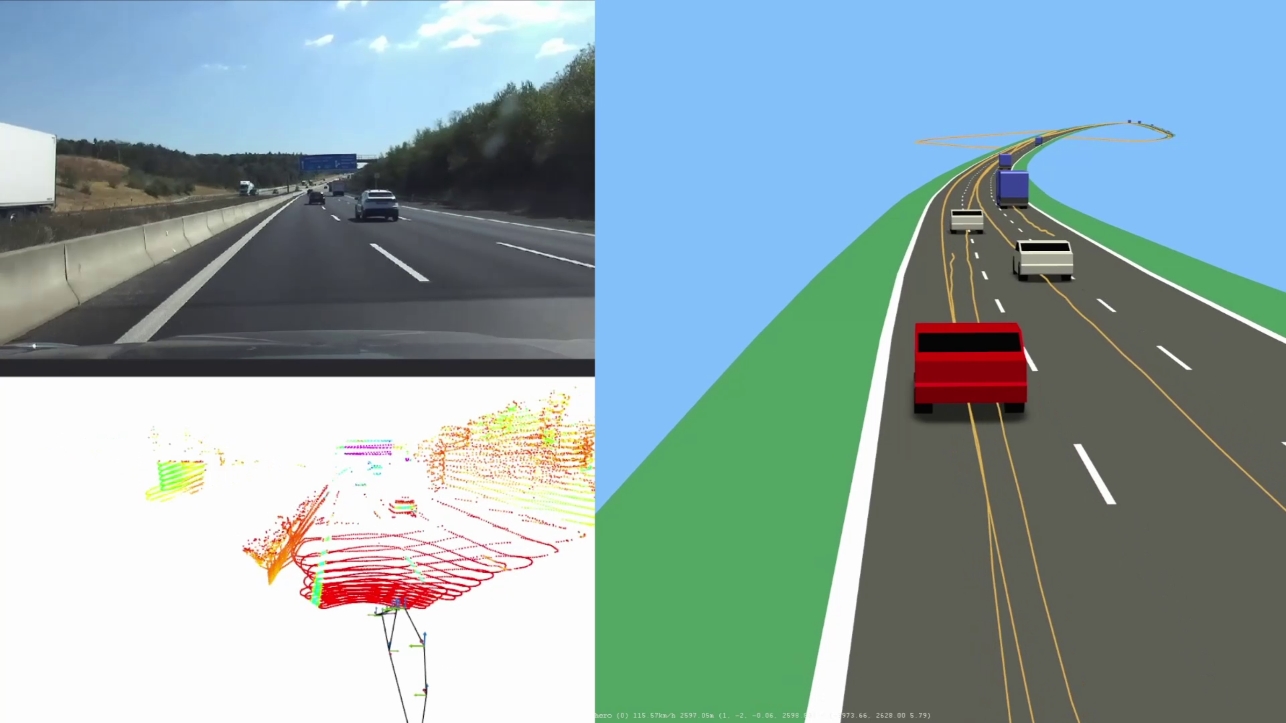}}\hfill
	\subfloat[1][\label{fig:qual-examples-2}]
	{\includegraphics[width=0.675\columnwidth]{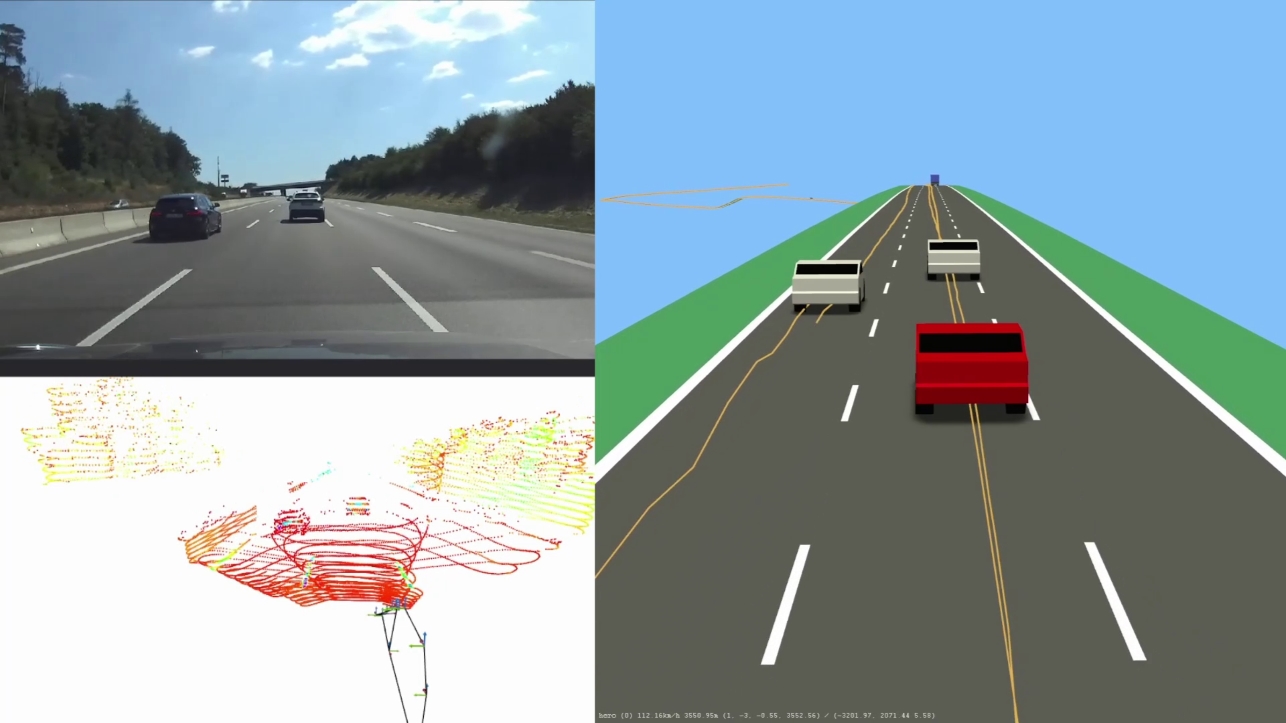}}\hfill
	\subfloat[1][\label{fig:qual-examples-3}]
	{\includegraphics[width=0.675\columnwidth]{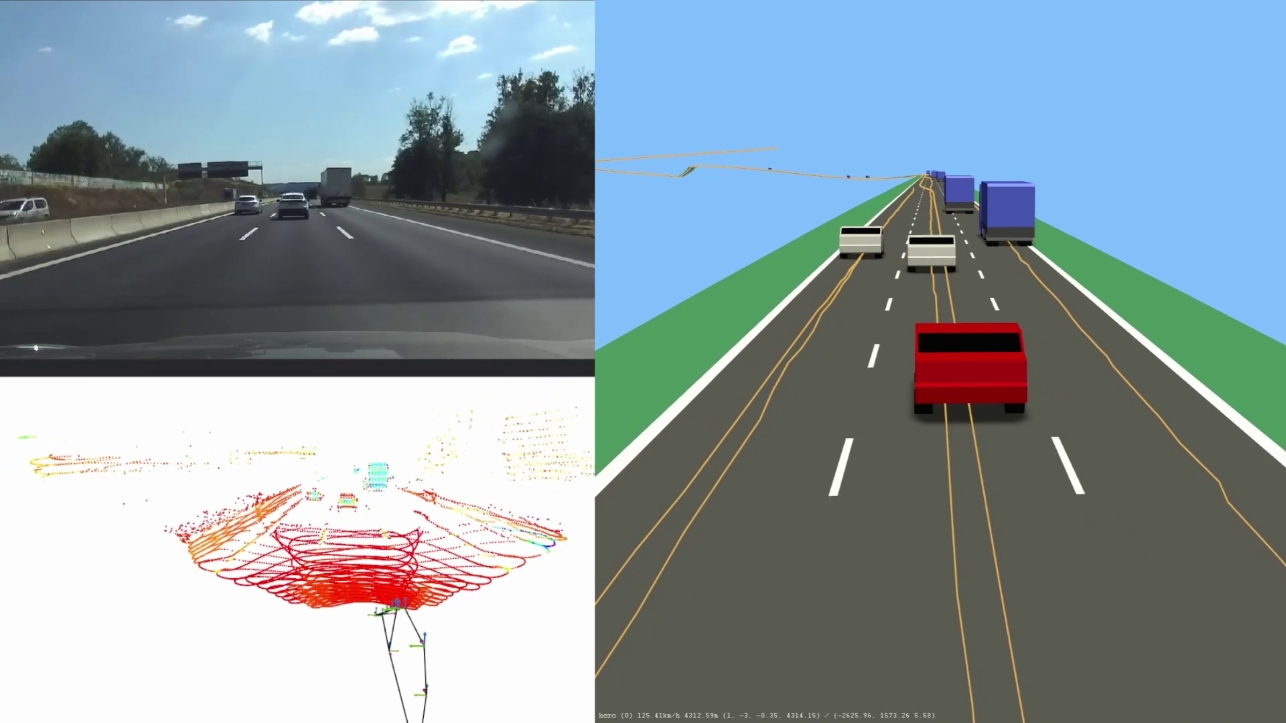}}\hfill
	\caption{Example excerpts of one full drive for qualitative comparison. The respective camera frame and LiDAR scan from the original recording are referenced in each image. The resulting OpenDRIVE is displayed as esmini \cite{esmini-environment-2023} rendering, including traffic participants from generated OpenSCENARIO\cite{asam-openscenario}.}
	\label{fig:qual-examples}
\end{figure*}

\section{Result and Discussion}\label{sec:results-discussion}

The state-of-the-art methods presented in Tab.~\ref{tab:sota-req-comparison} can reconstruct several hundred meters of road. In comparison, our method can successfully handle and reconstruct several kilometers of highway, as shown in Fig.~\ref{fig:qual-examples}. 

\subsection{Quantitative Analysis}

\begin{table}[t]
\caption{Evaluations of our resulting OpenDRIVE maps. For accuracy we compare against PEGASUS \cite{pegasus-project} HD maps and for reproducibility against our methodology.}
\centering
\begin{tabular}{@{}lcc@{}}
\toprule
                & \multicolumn{1}{l}{agst. PEGASUS HD} & \multicolumn{1}{l}{agst. self (ours)} \\ \midrule
RMSE            & 0.303 (m)                                 & 0.24 (m)                                   \\ \midrule
avg. distance   & 0.238 (m)                                 & 0.20 (m)                                   \\ \midrule
std. derivation $\sigma$ & 0.187 (m)                                 & 0.13 (m)                                   \\ \midrule
eval. length    & 44.8 (km)                                 & 30.6 (km)                                   \\\bottomrule
\end{tabular}

\label{tab:quant-eval}
\end{table}

The quantitative evaluation of map data depends on the relevant purpose and parameters, including the format or standard used, and is an ongoing research topic. 
The OpenDRIVE standard provides flexibility to express road layouts in different ways, leading to equivalence classes of roads of identical or near-identical physical layouts but considerably different parameters, thereby challenging the comparison of road models. For example, the relative position of \textit{reference line}, lanes, and georeference can be offset at multiple occasions.

Therefore, comparisons are performed on the resulting geometries of the considered OpenDRIVE maps rather than the parameters describing them. We note that within a local neighborhood, only lateral errors are measurable, with considered OpenDRIVE features being translation-invariant along the reference line except for locations where the lateral layout changes. Individual road marking locations, for example, are not mandatory in OpenDRIVE and are not included in the references since they can frequently change in practice due to road works.

Hence, we calculate a parametrization-invariant lateral distance between the roads through the global position of the reference lines in the reference map and ours. Comparison is made by sampling uniformly spaced points from each reference line. Between both lines, the nearest neighbors and the distance between them are calculated. Similar to \cite{chiang_automated_2022}, we evaluated the Root Mean Square Error (RMSE), average distance, and standard deviation ($\sigma$).

As the methodologies presented in Tab.~\ref{tab:sota-req-comparison} are vehicle-dependent, we evaluate against HD maps from the PEGASUS research project \cite{pegasus-project}. As input data, we use four different recordings, taken over the course of 8 weeks, following the same part of the German highway A8. The drives contain a total of 44.8 km recorded, including 31 cut-in scenarios and 29 ego vehicle lane changes, with an average speed of 116 km/h. Recording conditions range from sunny to light rain, with mid to high traffic volume, with speed profiles from 80 to 140 km/h.

Comparing the resulting 44.8 km reconstruction against PEGASUS HD maps yields an average distance of 0.238 m, an $\sigma$ of 0.187 m, and an RMSE of 0.303 m.
According to the Taiwan HD map standards \cite{taiwan-hd-map-standard-2020}, the 3D map accuracy should be lower than 30 cm. As shown in Tab.~\ref{tab:quant-eval}, the results of our proposed methodology satisfy these requirements.

A different reference is the reproducibility of results in repeated drives under different conditions described above. Thus, we choose one recording randomly as a reference map and evaluate the remaining three against it. 
We achieve a combined average distance of 0.20 m, with an $\sigma$ of 0.13 m and RMSE of 0.24 m. 
The measured offsets hereby originate from a multitude of factors. As initially stated, we only introduce moderate requirements on calibration and accuracy. Hence calibration offsets and GNSS offsets affect the results as well as 
the alignment of geo-references, and the curve fitting of the reference line.
Despite these offsets, the results show that our proposed method achieves reproducible maps across different traffic and driving situations. 

Regarding the estimated lane numbers, our approach reconstructs a total number of three lanes along the highway section, which is consistent with the true number of lanes, disregarding entry and exit lanes.

To provide a metric for the lateral accuracy of our approach, we evaluate the estimated lane width from Sec.~\ref{subsec:calc-ref-line}. Therefore, the two most left lanes are used, as these have similar lane width. Evaluated over the total of 44.8 km, our method yields an average lane width of 3.56 m, with an $\sigma$ of 0.22~m. Governmental regulations stipulate a lane width of 3.5 m with a dashed marker width of 0.15 m and 0.30 m for solid marks on the left road boundary\cite{frank-leitfaden-2017}.

\subsection{Qualitative Analysis}

For the qualitative analysis, we generate an additional OpenSCENARIO \cite{asam-openscenario} file and reference it to the OpenDRIVE. For the object detection of the traffic participants, we use the series sensors.

Figure~\ref{fig:qual-examples} demonstrates a set of frames from the corresponding esmini \cite{esmini-environment-2023} rendering. 
The full video is available on our data exchange
\footnote{\url{https://dataexchange.porsche-engineering.de/wl/?id=wiMyDDwFZUnb1CUokIuchMGuLnfgPuYy} \\Password: IEEE-ITSC23}.
The examples demonstrate the correct placement of the ego vehicle, as well as other traffic participants. Even in complex situations with a high number of vehicles, the generated road representation shows a close-to-real shape, with observed manoeuvrers, such as cut-in and cut-out, replicated correctly in the simulation. This is also achieved in cases where the ego vehicle executes a lane change or is on a non-center lane. Furthermore, not only the position on the lane shows high accuracy, but also the position within each lane. For example, in Fig.~\ref{fig:qual-examples-2}, \ref{fig:qual-examples-3}, the target vehicle is on the far right and left side of the lane, respectively, which is also closely reconstructed in the simulation.

As described, all vehicles in the simulation are placed in reference to the GNSS with an accuracy of $\pm$5 cm. Hereby we conclude that the road has a high degree of realism. Even in situations where the GNSS has a higher offset, e.g., through loss of GNSS signal, our method can correctly infer the road structure and display a smooth and homogeneous transition between geometric elements.

\section{Conclusion and Future Work}
Based on the importance of road networks for the evaluation of (highly) automated driving functions, research on their creation has seen considerable interest over the last few years.

In this paper, we proposed an algorithm that can automatically generate high-resolution road representations directly from real-world test drives. Leveraging sparse LiDAR data, the algorithm involves lane marking extraction, 3D lane line generation, and direct export into OpenDRIVE standard. Without relying on the prior road or odometry information, the algorithm is independent of the LiDAR sensor and allows accumulation from multiple sources.  
Furthermore, the algorithm's results fulfill the requirements for HD map standards and driving function simulation, which could be shown through quantitative and qualitative metrics. Simulated test drives show the accurate placement of traffic participants close to reality, even at ego lane changes.

Entry or exit lanes represent avenues for further research. We successfully integrated a classification in different tests by leveraging the distance between lane markings. As these are normally different from straight lanes, we were able to integrate them into the OpenDRIVE. Although, problems arise when the ego vehicle only partly records these special lanes. Similar situations currently occur with adaptively rising or lowering lane numbers.
Another point of interest for future research is the adaption of different sensors. As our method, by Sec.~\ref{subsec:line-marker-extrac}, only relies on a 2D or 3D representation of line markers, an adaption onto other sensors technologies, like cameras, is expected to be possible.

\bibliographystyle{IEEEtran}
\bibliography{bibopendrive}

\end{document}